\documentclass[10pt,twocolumn,letterpaper]{article}

\usepackage{cvpr}              % To produce the CAMERA-READY version
\usepackage{adjustbox} % To produce the REVIEW version
\newcommand{\TODO}[1]{\textbf{\color{red}[TODO: #1]}}
\renewcommand{\TODO}[1]{}

\usepackage{microtype}

\renewcommand{\paragraph}[1]{\vspace{.5em}\noindent\textbf{#1.}}

\usepackage{xspace}

\usepackage{soul}
\setuldepth{foobar}

\definecolor{cvprblue}{rgb}{0.21,0.49,0.74}
\usepackage[pagebackref,breaklinks,colorlinks,allcolors=cvprblue]{hyperref}

\def\paperID{18177} % *** Enter the Paper ID here
\def\confName{CVPR}
\def\confYear{2026}

\title{Don't let the information slip away}

\author{Taozhe Li\\
University of Oklahoma\\
{\tt\small Taozhe.Li-1@ou.edu}
\and
Guansu Wang\\
University of Melbourne\\
{\tt\small guansuw@student.unimelb.edu.au}
\and
Bo Yu\\
University of Utah\\
{\tt\small bo.yu@ou.edu}
\and
Yiming Liu\\
University of Oklahoma\\
{\tt\small ymliu@ou.edu}
\and
Wei Sun$^{*}$\\
University of Oklahoma\\
{\tt\small wsun@ou.edu}
}
\begin{document}
\maketitle
\begin{abstract}
Real-time object detection has advanced rapidly in recent years. The YOLO series of detectors is among the most well-known CNN-based object detection models and cannot be overlooked. The latest version, YOLOv26, was recently released, while YOLOv12 achieved state-of-the-art (SOTA) performance with 55.2 mAP on the COCO val2017 dataset. Meanwhile, transformer-based object detection models, also known as DEtection TRansformer (DETR), have demonstrated impressive performance. RT-DETR is an outstanding model that outperformed the YOLO series in both speed and accuracy when it was released. Its successor, RT-DETRv2, achieved 53.4 mAP on the COCO val2017 dataset. However, despite their remarkable performance, all these models let information to slip away. They primarily focus on the features of foreground objects while neglecting the contextual information provided by the background. We believe that background information can significantly aid object detection tasks. For example, cars are more likely to appear on roads rather than in offices, while wild animals are more likely to be found in forests or remote areas rather than on busy streets. To address this gap, we propose an object detection model called Association DETR, which achieves state-of-the-art results compared to other object detection models on the COCO val2017 dataset.
\end{abstract}

\begin{figure*}
  \centering
   \includegraphics[width=0.95\textwidth, height=0.35\textwidth]{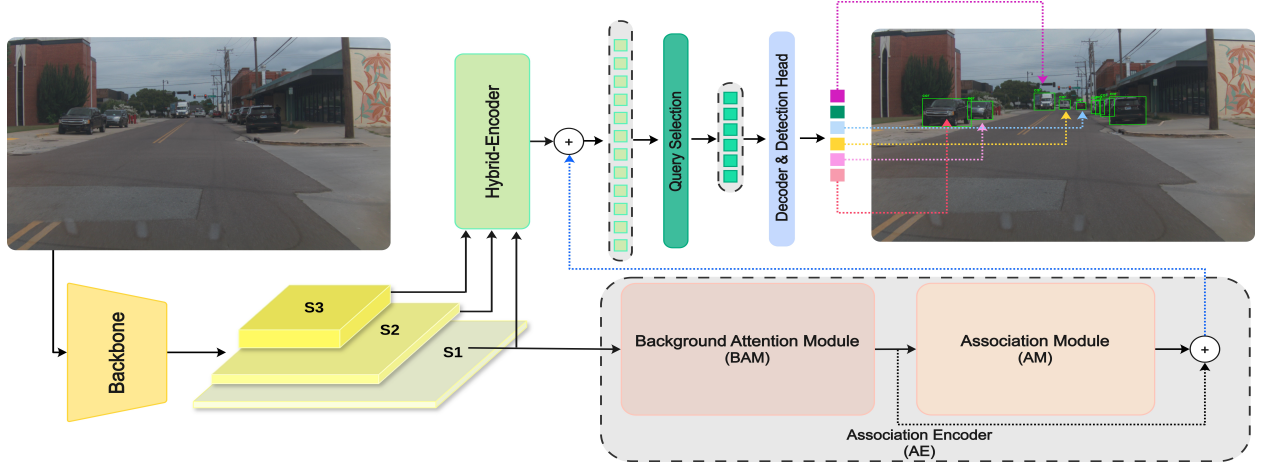}
   \caption{Association DETR Overview. The input image is first fed into the backbone network. The shallowest image feature is denoted as $S_1$, the second shallowest as $S_2$, and the deepest as $S_3$. we feed the shallowest feature $S_1$ into the Background Attention Module, which is designed to capture background information. Also in Figure \ref{background-attention}. We visualize some sample for better understanding the function of the Background Attention Module. And The Features $S_1$, $S_2$, and $S_3$ are fed into the Hybrid Encoder for both intra-feature and inter-feature enhancement. The output $F_b$ of the Background Attention Module is then fed into the Association Module, which performs feature enhancement related to background information. After that, the output of the Association Module, $F_a$ performs an addition operation with $F_b$. Additionally, Feature $F_b$ is added to $F_{3}$ as $F_{\hat{3}}$, the $F_{3}$ refers to the output of the Hybrid Encoder corresponding to the input $S_3$. Finally, the features $F_1$, $F_2$, and $F_{\hat{3}}$ undergo query selection and are passed into the Decoder and Detection Head to predict object bounding boxes and classes.}
   \label{Overview}
\end{figure*}    
\section{Introduction}
\label{sec:intro}

% \begin{figure*}[htbp]
%     \centering
%     \begin{minipage}[b]{0.25\textwidth}
%         \centering
%         \includegraphics[width=\linewidth, height=\linewidth]{sec/Figures/Spation_Attention_Module.jpg}
%     \end{minipage}
%     \hspace{0.1em} % move second image slightly left
%     \begin{minipage}[b]{0.25\textwidth}
%         \centering
% \includegraphics[width=2\linewidth, height=\linewidth]{sec/Figures/RFCVAMConvBlock.jpg}
%     \end{minipage}
%     \vspace{0.8em}
%     \caption{The Comma 3X device. Front side (top left) includes the touchscreen and driver-monitoring wide-angle camera; rear side (top right) includes a narrow-angle long-range camera and a wide-angle short-range camera. Bottom shows the custom car harness.}
%     \label{fig:Comma3x}
% \end{figure*}

Artificial intelligence has made rapid progress in recent years and has been applied in various fields, including but not limited to Simultaneous Localization and Mapping (SLAM) \cite{mlp-slam}, Autonomous Driving, Large Language Model \cite{llm1, Feng2026NoisyValid}, Medical Image Analysis, Behavioral Monitoring, Privacy\cite{Feng1, Feng2, privacy2} and more. And object detection remains one of the most active and popular research areas among these fields. The mainstream object detection models consist of two branches: one is CNN-based architectures, and the most famous one in this branch is the YOLO series detectors \cite{yolo11_ultralytics, yolo12, yolov8_ultralytics}. The other branch comprises end-to-end Transformer-based detectors \cite{carion2020endtoendobjectdetectiontransformers}.

The YOLO series detectors  primarily focus on trade-off between speed and accuracy, as not all devices—especially those used in autonomous driving possess powerful computational capabilities. This indicates that the performance of YOLO detectors is generally inferior to that of end-to-end Transformer-based detectors. The YOLO series detector recently released its 12th version \cite{yolo12}. However, it just shows limited performance improvement compared to its previous version, YOLOv11 (1.0 mAP in the N, S and M size models, 0.3 mAP in the L size model, and 0.5 mAP in the X size model).

The end to end Transformer based detector (DETRs) have received extensive attention recently. The latest architecture in DETRs named DEIMv2 \cite{huang2025deimv2}, it has achieved state-of-the-art (SOTA) performance with a 57.80 mAP on the COCO\cite{coco2017} val 2017 dataset, which is remarkable. However, the best-performing version of DEIMv2(DEIMv2-X) is built by almost 50 million parameters, making it extremely difficult to deploy on the onboard devices that typically lack strong computational power, which is very common in the autonomous driving field. This means that the speed is critical for object detectors.

Due to the high performance and streamlined architecture of DETRs, as well as the urgent need for speed, some researchers have concentrated on real-time DETRs. The work RT-DETR \cite{rt-detr} is a DETRs that cannot be overlooked. It outperforms the YOLO series detectors in both speed and accuracy at the time it was proposed, achieving state-of-the-art performance.

However, although the aforementioned YOLO series detectors and DETRs have achieved remarkable performance, they still allow valuable information to slip away. Most of these models focus primarily on foreground information while neglecting the background. The study \cite{DBLP:journals/corr/abs-2006-09994} demonstrates that background information can be extremely useful for computer vision tasks. In that work, even when the foreground object is removed from the image, the classification model can still achieve around 50\% accuracy, compared to only 11.11\% accuracy from random guessing. From our perspective, the way background information aids computer vision tasks, such as object detection or classification, is similar to the associative ability humans possess. More specifically, when we see a photo taken indoors, if we need to guess what objects are present, we are likely to guess a person, a sofa, or a clock on the wall, and it is highly unlikely we would guess a car or a traffic light. To address the issue of losing background information, we propose an Association DETR equipped with an Association Encoder. The contributions of our work are summarized as follows:
\begin{enumerate}
    \item We propose Association DETR, a model that captures both background and foreground information, achieving state-of-the-art performance on the COCO 2017 dataset with a 54.6 mAP.
    \item The Association Encoder is a plug-in module with only 3 million parameters that can be easily installed on any existing DETR model to boost its performance.
\end{enumerate}

% \begin{figure*}
%   \centering
%   \begin{subfigure}{0.68\linewidth}
%     \fbox{\rule{0pt}{2in} \rule{.9\linewidth}{0pt}}
%     \caption{An example of a attention.}
%     \label{fig:short-a}
%   \end{subfigure}
%   \hfill
%   \begin{subfigure}{0.28\linewidth}
%     \fbox{\rule{0pt}{2in} \rule{.9\linewidth}{0pt}}
%     \caption{Another example of a attention}
%     \label{fig:short-b}
%   \end{subfigure}
%   \caption{}
%   \label{fig:short}
% \end{figure*}

\begin{table*}
  \caption{\textbf{Comparison with SOTA}. All models are trained on the COCO Dataset. And the FPS is reported on T4 GPU with TensorRT. For Evaluation, all input size are fixed on 640$\times$640, Our Association DETR-R34 achieves 54.60 in AP$^{val}$, 71.6 in AP$^{val}_{50}$ and 153 FPS.Meanwhile, our Association DETR-R50 achieves 55.7 in AP$^{val}$, 74.0 in AP$^{val}_{50}$ and 104 FPS. Both of them outperform other YOLO detectors and DETR models with similar scale.}
  \label{sota_comparison_table}
  \centering
  \begin{tabular}{c@{}c c c c c c c c c c }
    \toprule
    Model & Backbone & \#Epochs & \#Params (M) & \#FPS$_{bs=1}$ & AP$^{val}$ & AP$^{val}_{50}$ \\
    \midrule
    YOLOv10-M & - & 300 & 15.4  & 210 & 51.1 & 68.1  \\
    YOLOv10-L & - & 300 & 24.4  & 137 & 53.2 & 70.1  \\
    YOLOv10-X & - & 300 & 29.5  & 93 & 54.4 & 71.6  \\
    YOLOv11-M & - & 300 & 20.1  & 212 & 51.5 & 67.9  \\
    YOLOv11-L & - & 300 & 25.3  & 161 & 53.4 & 70.1  \\
    YOLOv12-M & - & 300 & 20.2  & 206 & 52.5 & 69.5  \\
    YOLOv12-L & - & 300 & 26.4  & 148 & 53.7 & 70.6  \\
    RT-DETR & R34 & 72 & 31  & 161 & 48.9 & 66.8  \\
    RT-DETRv2-M & R34 & 72 & 31 & 161 & 49.9 & 67.5  \\
    Association-DETR-R34 (Ours) & R34 & 72 & 34.1 & 153 & 
    \textbf{54.6} & \textbf{71.6} \\
    \midrule
    DETR-DC5 & R50 & 500 & 41  & - & 43.3 & 63.1   \\
    DETR-DC5 & R101 & 500 & 60  & - & 44.9 & 64.7   \\
    Deformable-DETR  & R50 & 108 & 44  & - & 45.1 & 65.4  \\
    RT-DETR & R50 & 72 & 42  & 108 & 53.1 & 71.3   \\
    RT-DETR & R101 & 72 & 76 & 74 & 54.3 & 72.7  \\
    YOLOv11-X & - & 300 & 56.9  & 88 & 54.7 & 71.6 \\
    YOLOv12-X & - & 300 & 59.1  & 85 & 55.2 &  72.2 \\
    RT-DETRv2-L & R50 & 72 & 42  & 145 & 53.4 & 71.6   \\
    RT-DETRv2-X & R101 & 72 & 76  &  74 & 54.3 & 72.8 \\
    Association-DETR-R50 (Ours) & R50 & 72 & 45.1  & 104 & \textbf{55.7} & \textbf{74.0}  \\
    \midrule
    \bottomrule
  \end{tabular}
\end{table*}

\begin{table*}
  \caption{\textbf{Effectiveness}. We integrate our Association Encoder(AE) into various models, including RT-DETRv34, RT-DETR-R50, RT-DETRv2-R50, DETRv2-R34, DETR and Deformable DETR.Specifically, for the RT-DETR-R34 model, it increases AP$^{val}$ by 5.7 and AP$^{val}_{50}$ by 4.8, while reducing FPS by less than 5.7\%. Similarly, it improves AP$^{val}$ by 2.6 and AP$^{val}_{50}$ by 2.7 for RT-DETR-R50.  Moreover, 2.5 in AP$^{val}$ and 2.7 in AP$^{val}_{50}$ for RT-DETRv2-M; 0.3 in AP$^{val}$ and  0.6 in AP$^{val}_{50}$ for RT-DETRv2-L with R50 as backbone; DETR-R50, it increases 2.7 in AP$^{val}$ and 3.6 in AP$^{val}_{50}$. We need to noticed that it even outperforms than DETR-R101(base). Moreover, 2.6 in AP$^{val}$ and 3.2 in AP$^{val}_{50}$ for Deformable DETR. All these models outperform their respective baselines after integrating our Association Encoder (AE). Notably, the Association-DETR-R34 and Association-DETR-R50 models achieve state-of-the-art performance compared to other models.}
\label{effectiveness_table}
  \centering
  \begin{tabular}{c@{}c c c c c c c c c c c}
    \toprule
    Model & Backbone & \#Epochs & \#Params (M) & \#FPS$_{bs=1}$ & AP$^{val}$ & AP$^{val}_{50}$ \\
    \midrule
    RT-DETR-R34 (base) & R34 & 72 & 31 &  161 & 48.9 & 66.8  \\
    AE + DETR-R34 (Ours) & R34 & 72 & 34.1  & 153 & \textbf{54.6($\uparrow$\textbf{5.7})} & \textbf{71.6}($\uparrow$\textbf{4.8}) \\
    RT-DETR-R50 (base) & R50 & 72 & 42  & 108 & 53.1 & 71.3  \\
    AE + DETR-R50 (Ours) & R50 & 72 & 45.1 & 104 & \textbf{55.7}($\uparrow$\textbf{2.6}) & \textbf{74.0}($\uparrow$\textbf{2.7})  \\
    RT-DETRv2-M (base) & R34 & 72 & 31  & 161 & 49.9 & 67.5  \\
    AE + RT-DETRv2-M & R34 & 72 & 34.1  & 153 & \textbf{52.4}($\uparrow$\textbf{2.5}) & \textbf{70.2}($\uparrow$\textbf{2.7})\\
    RT-DETRv2-L (base) & R50 & 72 & 42  & 145 & 53.4 & 71.6 \\ 
    AE + RT-DETRv2-L & R50 & 72 & 45.1  & 137 & \textbf{53.7}($\uparrow$\textbf{0.3}) & \textbf{72.1}($\uparrow$\textbf{0.6}) \\
    DETR-R50 (base)  & R50 & 500 & 41  & - & 43.3 & 63.1 \\ 
    DETR-R101 (base) & R101 & 500 & 60  & - & 44.9 & 64.7\\
    AE + DETR & R50 & 500 & 63.1  & - & \textbf{46.0}($\uparrow$\textbf{2.7}) & \textbf{66.7}($\uparrow$\textbf{3.6}) \\ 
    Deformable-DETR(base) & R50 & 108 & 44  & - & 45.1 & 65.4\\
    AE + Deformable DETR & R50 & 108 & 47.1 & - & \textbf{47.7}($\uparrow$\textbf{2.6}) & \textbf{68.6}($\uparrow$\textbf{3.2})\\
    \bottomrule
  \end{tabular}
\end{table*}

\section{Related Work}
%------------------------------------------------------------------------
\subsection{Real-time Object Detectors}
YOLOv1 \cite{yolov1} is the first CNN-based, one-stage object detector to achieve true real-time object detection. After years of continuous development, the YOLO series of detectors has significantly advanced the field.

Outperforming other one-stage object detectors \cite{girshick2015fast, liu2016ssd}, YOLO has become synonymous with real-time object detection and has been applied in various fields. Moreover, the YOLO series of detectors can be classified into two categories:
Anchor-based \cite{yolov1, redmon2018yolov3, bochkovskiy2020yolov4, yolov5} and anchor-free \cite{yolov8_ultralytics, THU-MIGyolov10, yolo11_ultralytics, yolo12},  methods.But only anchor-free YOLO series detectors achieve a reasonable trade-off between speed and accuracy, as demonstrated in \cite{rt-detr}.

Moreover, the YOLO series of detectors has evolved rapidly, with the latest version announced as YOLO26 \cite{sapkota2025ultralytics}. Additionally, the X version of YOLOv10 \cite{THU-MIGyolov10}, YOLOv11 \cite{yolo11_ultralytics}, and YOLOv12 \cite{yolo12} achieve impressive performance on the COCO 2017 dataset \cite{coco2017}, with mean Average Precision (mAP) scores of 54.4, 54.7, and 55.2, respectively. 

However, all these models focus exclusively on foreground information, neglecting background information that could potentially further enhance model performance.

\subsection{End-to-End Object Detectors}
End-to-end object detectors based on Transformers, also known as Detection Transformers, were first proposed by Carion et al. \cite{carion2020endtoendobjectdetectiontransformers}. These models have attracted extensive attention since it has been posted. Compared to the YOLO series detectors, they possess distinctive features, a streamlined structure, and significant potential. However, they also face several challenges, including slow training convergence, high computational cost, and difficult-to-optimize queries. Many DETR variants have been developed to address these issues.
For solving the problem of slow convergence time, the Deformable-DETR \cite{zhu2020deformable} introduce utilizing multi-scale features and enhancing the efficiency of the attention mechanism. And the SAM-DETR \cite{zhang2022accelerating} improves convergence speed through semantic information matching. Moreover, DAB-DETR \cite{liu2022dab} and DN-DETR \cite{li2022dn} further enhance performance by introducing iterative refinement schemes and denoising training. 

To reduce computational cost, Group-DETR \cite{chen2022group, chen2023group} employs group-wise one-to-many assignment. RT-DETR \cite{rt-detr} proposes a hybrid-efficient encoder and an improved query selection method to further reduce computation. The upgraded version, RT-DETRv2 \cite{rtdetrv2}, introduces selective multi-scale sampling to better handle objects of varying sizes. Additionally, Efficient DETR \cite{yao2021efficient} and Sparse DETR \cite{roh2021sparse} reduces computational cost by decreasing the number of encoder and decoder layers or the number of updated queries. And the Lite DETR \cite{li2023lite} improves encoder efficiency by lowering the update frequency of low-level features.

To decreasing the optimization difficulty of queries.The features are processed in an interleaved manner is utilized in Conditional DETR \cite{meng2021conditional} and Anchor DETR \cite{wang2022anchor}. Also, work \cite{zhu2020deformable} proposes a query selection method that specialized for two-stage DETR.

However, all the aforementioned DETRs and YOLO series detectors focus exclusively on foreground information and executing inter and intra enhancement on it, but none of them explicitly leverage background information. Our Association-DETR bridges this gap, it acheves SOTA performance on the COCO 2017 dataset with a 54.6 mAP as presented in Table \ref{sota_comparison_table}. Also, according to the experimental results in Table \ref{effectiveness_table}, the proposed light Association Encoder can be integrated into any existing DETR model to effectively boost its performance while minimal impact on inference time.

%-------------------------------------------------------------------------

\subsection{Spatial Attention \& Channel Attention}
Spatial attention refers to the mechanism by which cognitive and computational systems selectively focus on specific spatial regions to enhance the processing of relevant visual information. It has been widely used in computer vision tasks to improve model performance. For example, the Convolutional Block Attention Module (CBAM) \cite{woo2018cbam} is a notable advancement in this field. It enhances the representational capacity of convolutional neural networks (CNNs) by incorporating both channel and spatial attention mechanisms. By strengthening the network's representational power, CBAM improves classification performance on the ImageNet-1K dataset \cite{5206848}. Additionally, the Coordinate Attention (CA) \cite{hou2021coordinate} incorporates positional information into channel attention, further boosting model performance in classification and object detection tasks. Moreover, compared to CBAM and CA, the proposed Receptive-Field Attention (RFA) in \cite{zhang2023rfaconv} not only focuses on spatial features but also addresses the issue of parameter sharing in convolutional kernels. RFA achieves better performance than CBAM and CA on both classification and object detection tasks. Furthermore, the combination of RFA and CBAM, termed RFCBAMConv, attains state-of-the-art performance on the ImageNet-1K classification task. Therefore, we will utilize RFCBAMConv in our subsequent Background Attention Module (BAM). Experimental results are presented in Table \ref{ablation-table} in Section \ref{sec:ablation}.

\begin{figure*}
\centering
\includegraphics[width=0.8\textwidth, height=0.3\textwidth]{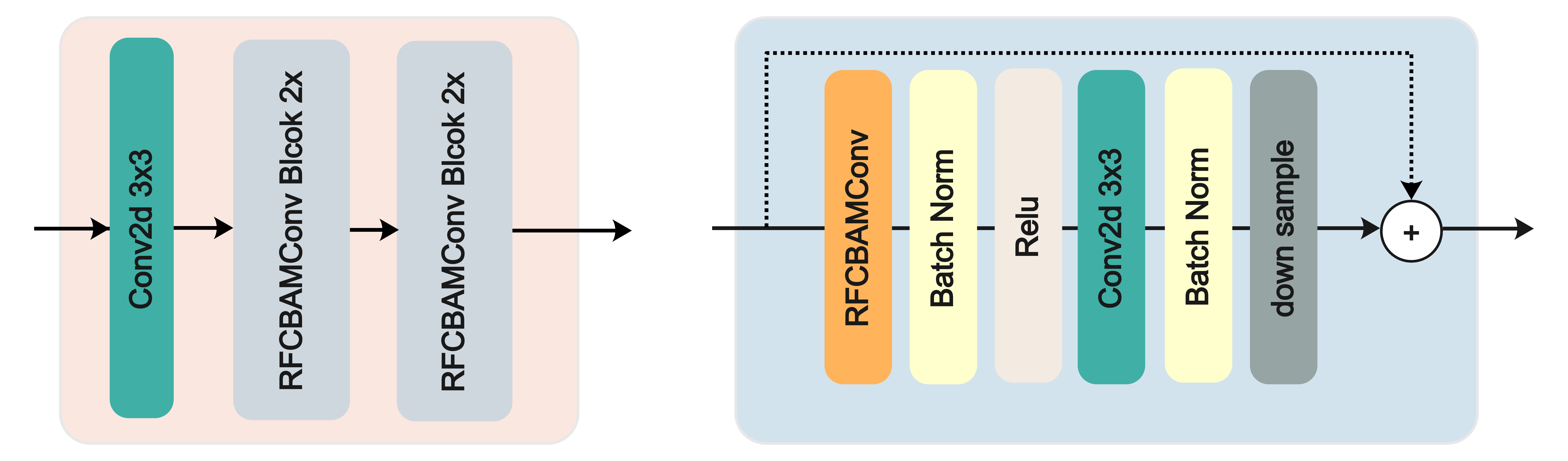}
\caption{Background Attention Module \& Single RFCBAMConv Block.On the left side is the structure of the Background Attention Module, and on the right side are the details of a single RFCBAMConv Block, which is located within the Background Association Module.}
\label{background-attention-module}
\end{figure*}
\section {Association DETR}
\begin{figure*}
  \centering
   \includegraphics[height=0.3\textwidth]{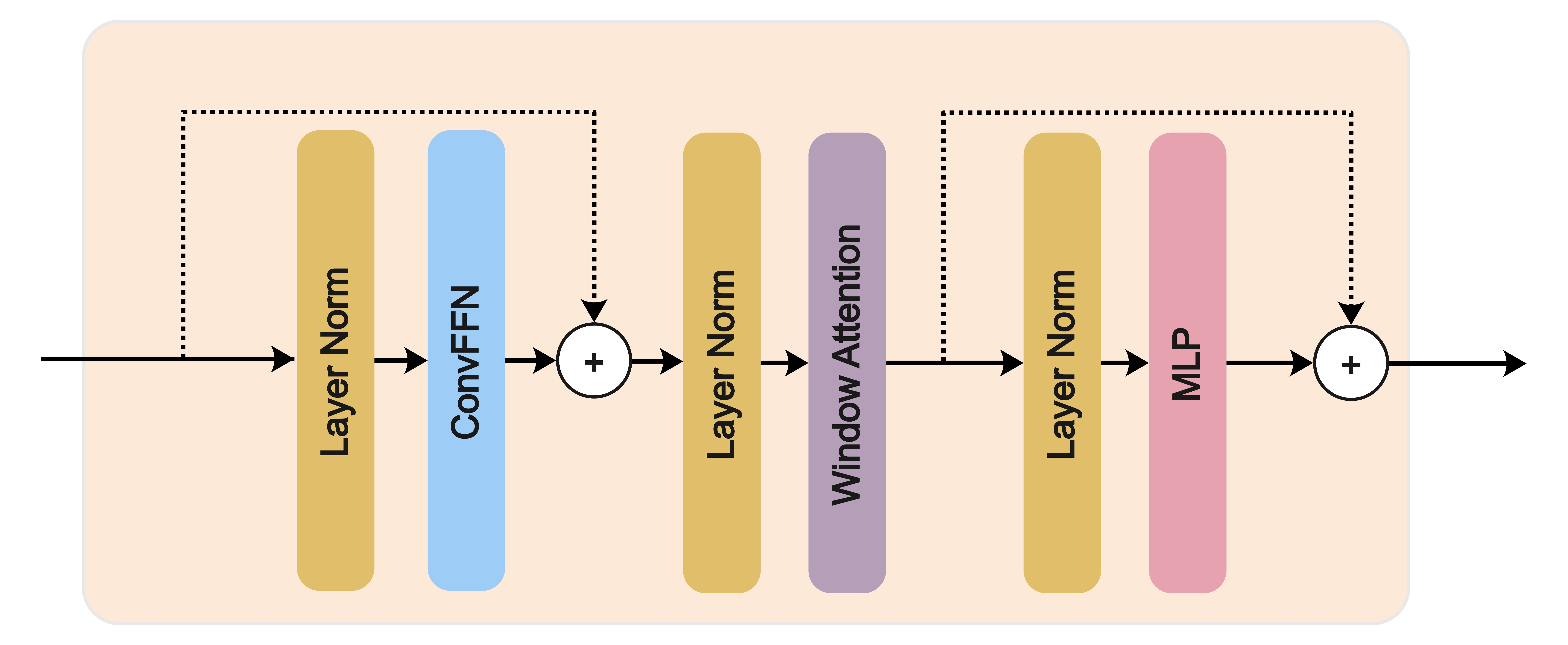}
   \caption{Association Module. We incorporate ConvFFN and Window Attention for trade off between performance and speed.}
   \label{association-module}
\end{figure*}
\subsection{Overview}
The overview of Association DETR is presented in Figure \ref{Overview}. Our model uses the powerful RT-DETR \cite{rt-detr} as the baseline, upon which we build a Association Encoder which consists of two modules: the Background Attention Module and the Association Module. Each component of these modules is carefully designed to achieve an optimal balance between speed and efficiency. The input image is first fed into the backbone network. For fair comparison with other models and also to ensure stable performance, we select ResNet-34 and ResNet-50 as our backbones. We extract multi-level features from the backbone for further processing. The shallowest image feature is denoted as $S_1$, the second shallowest as $S_2$, and the deepest as $S_3$. Utilizing multi-level features is a common technique in computer vision field, as it has been shown to improve model performance \cite{lin2017feature} in different computer vision tasks. Typically, shallow image features captured by neural network is mainly consists of basic information such as edges, corners, and textures, while deeper neural network captures high-level features, including semantic information, object parts, or even specific objects. For these reasons, we feed the shallowest feature $S_1$ into the Background Attention Module, which is designed to capture background information. Also in Figure \ref{background-attention}. We visualize some sample for better understanding the function of the Background Attention Module. The visualization is plotted by tool pytorch-grad-cam \cite{jacobgilpytorchcam} .And The Features $S_1$, $S_2$, and $S_3$ are fed into the Hybrid Encoder for both intra-feature and inter-feature enhancement. The output $F_b$ of the Background Attention Module is then fed into the Association Module, which performs feature enhancement related to background information. After that, the output of the Association Module, $F_a$ performs an addition operation with $F_b$. This design helps preventing the vanishing gradient problem and enriches the extracted background information. Additionally, Feature $F_b$ is added to $F_{3}$ as $F_{\hat{3}}$, the $F_{3}$ refers to the output of the Hybrid Encoder corresponding to the input $S_3$. This operation further enriches the original image feature and serves as another feature enhancement technique. Finally, the features $F_1$, $F_2$, and $F_{\hat{3}}$ undergo query selection and are passed into the Decoder and Detection Head to predict object bounding boxes and classes.
\subsection{Association Encoder}
The Association Encoder (AE) is light encoder we carefully elaborate with only 3.1 million parameters. It consists of two components: the Background Attention Module (BAM) and the Association Module (AM). These modules are designed for background feature extraction and intra-feature enhancement respectively. Furthermore, we find that the AE is an efficient plug-in module that can be integrated into any DETR detector to improve the model's performance, as demonstrated by the experimental results in Table \ref{effectiveness_table}. The details of each module are described below.
\subsubsection{Background Attention Module}
\begin{figure*}
\centering
\adjustbox{valign=m}{
\begin{minipage}[m]{0.09\textwidth}
  \centering
  \textbf{BG Attention}\\[-1mm]
\end{minipage}}%
\hfill
\adjustbox{valign=m}{
\begin{subfigure}[m]{0.22\textwidth}
  \centering
  \includegraphics[width=\linewidth]{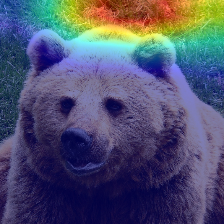}
\end{subfigure}}%
\hfill
\adjustbox{valign=m}{
\begin{subfigure}[m]{0.22\textwidth}
  \centering
  \includegraphics[width=\linewidth]{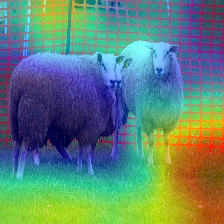}
\end{subfigure}}%
\hfill
\adjustbox{valign=m}{
\begin{subfigure}[m]{0.22\textwidth}
  \centering
\includegraphics[width=\linewidth]{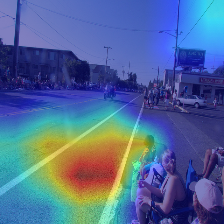}
\end{subfigure}}%
\hfill
\adjustbox{valign=m}{
\begin{subfigure}[m]{0.22\textwidth}
    \centering
\includegraphics[width=\linewidth]{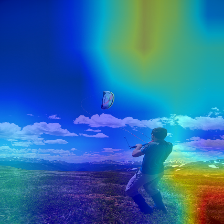}
\end{subfigure}}%

%==========================
\centering
\adjustbox{valign=m}{
\begin{minipage}[m]{0.09\textwidth}
  \centering
  \textbf{Image}\\[-1mm]
\end{minipage}}%
\hfill
\adjustbox{valign=m}{
\begin{subfigure}[m]{0.22\textwidth}
  \centering
  \includegraphics[width=\linewidth, height=\linewidth]{sec/Figures/bear.jpg}
\end{subfigure}}%
\hfill
\adjustbox{valign=m}{
\begin{subfigure}[m]{0.22\textwidth}
  \centering
  \includegraphics[width=\linewidth, height=\linewidth]{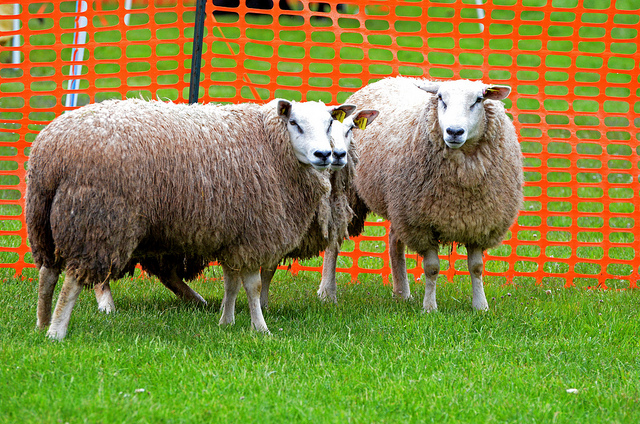}
\end{subfigure}}%
\hfill
\adjustbox{valign=m}{
\begin{subfigure}[m]{0.22\textwidth}
  \centering
\includegraphics[width=\linewidth, height=\linewidth]{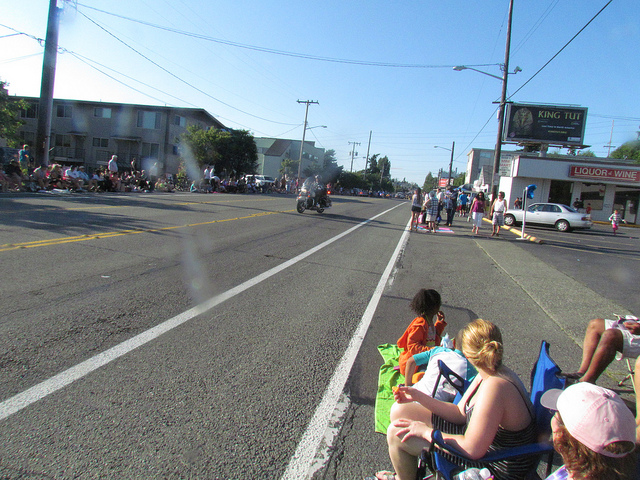}
\end{subfigure}}%
\hfill
\adjustbox{valign=m}{
\begin{subfigure}[m]{0.22\textwidth}
    \centering
\includegraphics[width=\linewidth, height=\linewidth]{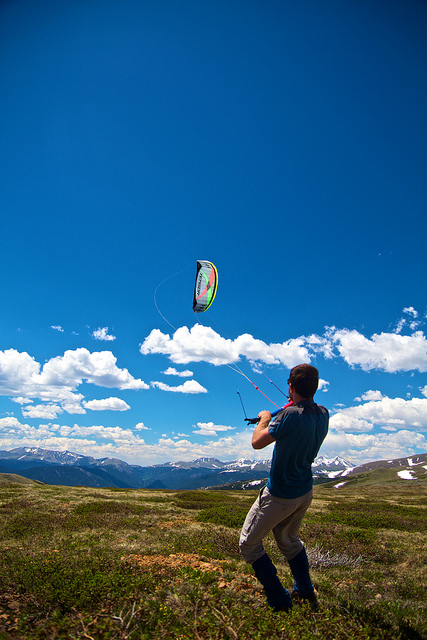}
\end{subfigure}}%
\caption{Visualization of Background Attention. BG refers to background. The figure is plotted by pytorch-grad-cam. The BAM (Background Attention Module) effectively captures background information across various scenarios. In the first image featuring bears, it successfully identifies the grass behind them. In the second image, it accurately detects grass and even fencing, despite fencing is not included in our training categories. Furthermore, in the third and fourth images, the module correctly identifies the road, sky, and grass. }
\label{background-attention}
\end{figure*}

The background attention module is designed to extract background information effectively. The detailed structure of our proposed background attention module is illustrated in Figure \ref{background-attention-module}. We carefully designed each component to optimize both performance and speed. The detailed architecture of the RFCBAMConv block is shown on the right side of Figure \ref{background-attention-module}. The RFCBAMConv is derived from the RFA (Receptive-Field Attention) method \cite{zhang2023rfaconv}. It combines RFA and CBAM, achieving state-of-the-art (SOTA) performance in classification tasks. This also demonstrates that RFCBAMConv has a strong capability for extracting image features. We apply this RFCBAMConv in the BM Module for better extract spatial image features, specifically background information. Additionally, some readers may notice that the structure of this BAM resembles ResNet blocks. This similarity is intentional because we pre-train the BAM before applying it to object detection tasks. The goal of this module is to extract background information efficiently while using fewer parameters. The simplest approach is to build a whole model similar to ResNet, maintaining the same number of layers and parameters but replacing all convolutional layers with RFCBAMConv. However, this results in a large number of parameters; even the smallest model contains 11 million parameters. To address this issue, we share the first two blocks with our backbone (ResNet has a total of four blocks) and train only the two blocks which located in BAM for background information extraction. This approach is effective and requires fewer parameters compared to the previously mentioned method(It is one-quarter of the number parameters of previous method). It is pre-trained on the Stanford Background Dataset \cite{gould2009decomposing} as a classification task. It includes a total of 9 background categories \cite{gould2009decomposing}, such as grass, road, tree, and sky. Then, the global average pooling layer and prediction head are removed before integrating it into any object detection model. This method significantly reduces the number of parameters required by the model and increases inference speed. If we were to use a complete ResNet-like model for the background attention module, it would increase the number of parameters by approximately five times compared to our method, resulting in much slower inference. The visualization results of background attention module that tested on some random images are presented in Figure \ref{background-attention}. As shown, the module effectively captures background information across various scenarios. In the first image featuring bears, it successfully identifies the grass behind them. In the second image, it accurately detects grass and even fencing, despite fencing is not included in our training categories. Furthermore, in the third and fourth images, the module correctly identifies the road, sky, and grass.

\subsubsection{Association Module}
The AM is designed to convert the extracted background information from the background-attention module into association information relevant to object detection. Essentially, it functions as a feature enhancement module from our perspective. Its structure is illustrated in Figure \ref{association-module}. To balance performance and speed, we incorporate ConvFFN \cite{tang2025convffn} and Window Attention \cite{zhang2022windowattention} within this module. ConvFFN serves as a feature extraction component that is significantly more efficient than self-attention. We made minor modifications to adapt it for the AM in the object detection task, as it was originally developed for super-resolution tasks. Additionally, Window Attention is also a optimization between performance and speed. It achieves comparable performance to multi-head attention but with a time complexity of only $O(n \times w)$, whereas the latter has a quadratic time complexity of $O(n^{2})$. Here, $n$ represents the number of image patches, and $w$ denotes the fixed number of windows. For evaluating the importance of this module, we conducted ablation experiments presented in Section \ref{sec:ablation}. And the experimental result is presented in Table \ref{ablation-table}. The module contributes an improvement of 1.3 mAP while adding only 0.7 million parameters. Furthermore, in the ablation study, we replaced this module with an EL (basic encoder layer) and found that although the EL has 0.4 million more parameters than the AM, it performs worse both individually and when combined with BAM.

\begin{table*}
  \caption{\textbf{Effectiveness}. We integrate our Association Encoder(AE) into various models, including RT-DETRv34, RT-DETR-R50, RT-DETRv2-R50, DETRv2-R34, DETR and Deformable DETR.Specifically, for the RT-DETR-R34 model, it increases AP$^{val}$ by 5.7 and AP$^{val}_{50}$ by 4.8, while reducing FPS by less than 5.7\%. Similarly, it improves AP$^{val}$ by 2.6 and AP$^{val}_{50}$ by 2.7 for RT-DETR-R50.  Moreover, 2.5 in AP$^{val}$ and 2.7 in AP$^{val}_{50}$ for RT-DETRv2-M; 0.3 in AP$^{val}$ and  0.6 in AP$^{val}_{50}$ for RT-DETRv2-L with R50 as backbone; DETR-R50, it increases 2.7 in AP$^{val}$ and 3.6 in AP$^{val}_{50}$. We need to noticed that it even outperforms than DETR-R101(base). Moreover, 2.6 in AP$^{val}$ and 3.2 in AP$^{val}_{50}$ for Deformable DETR. All these models outperform their respective baselines after integrating our Association Encoder (AE). Notably, the Association-DETR-R34 and Association-DETR-R50 models achieve state-of-the-art performance compared to other models.}
\label{effectiveness_table}
  \centering
  \begin{tabular}{c@{}c c c c c c c c c c c}
    \toprule
    Model & Backbone & \#Epochs & \#Params (M) & \#FPS$_{bs=1}$ & AP$^{val}$ & AP$^{val}_{50}$ \\
    \midrule
    RT-DETR-R34 (base) & R34 & 72 & 31 &  161 & 48.9 & 66.8  \\
    AE + DETR-R34 (Ours) & R34 & 72 & 34.1  & 153 & \textbf{54.6($\uparrow$\textbf{5.7})} & \textbf{71.6}($\uparrow$\textbf{4.8}) \\
    RT-DETR-R50 (base) & R50 & 72 & 42  & 108 & 53.1 & 71.3  \\
    AE + DETR-R50 (Ours) & R50 & 72 & 45.1 & 104 & \textbf{55.7}($\uparrow$\textbf{2.6}) & \textbf{74.0}($\uparrow$\textbf{2.7})  \\
    RT-DETRv2-M (base) & R34 & 72 & 31  & 161 & 49.9 & 67.5  \\
    AE + RT-DETRv2-M & R34 & 72 & 34.1  & 153 & \textbf{52.4}($\uparrow$\textbf{2.5}) & \textbf{70.2}($\uparrow$\textbf{2.7})\\
    RT-DETRv2-L (base) & R50 & 72 & 42  & 145 & 53.4 & 71.6 \\ 
    AE + RT-DETRv2-L & R50 & 72 & 45.1  & 137 & \textbf{53.7}($\uparrow$\textbf{0.3}) & \textbf{72.1}($\uparrow$\textbf{0.6}) \\
    DETR-R50 (base)  & R50 & 500 & 41  & - & 43.3 & 63.1 \\ 
    DETR-R101 (base) & R101 & 500 & 60  & - & 44.9 & 64.7\\
    AE + DETR & R50 & 500 & 63.1  & - & \textbf{46.0}($\uparrow$\textbf{2.7}) & \textbf{66.7}($\uparrow$\textbf{3.6}) \\ 
    Deformable-DETR(base) & R50 & 108 & 44  & - & 45.1 & 65.4\\
    AE + Deformable DETR & R50 & 108 & 47.1 & - & \textbf{47.7}($\uparrow$\textbf{2.6}) & \textbf{68.6}($\uparrow$\textbf{3.2})\\
    \bottomrule
  \end{tabular}
\end{table*}
% https://www.overleaf.com/project/697be46b58c35725e80ff227
\begin{table*}
  \caption{\textbf{Ablation}. All models are trained on the COCO 2017 dataset. For training, the number of epochs is fixed at 72 to ensure a fair comparison. For evaluation, all input sizes are fixed at 640×640. BAM, AM, and EL represent the Background Attention Module, Association Module, and Basic ViT Encoder Layer, respectively.}
 \label{ablation-table}
  \centering
  \begin{tabular}{c@{}c c c c c c}
    \toprule
    Model & BAM & AM & EL & \#Params (M) & AP$^{val}$ & AP$^{val}_{50}$ \\
    \midrule
    RT-DETR-R34 &  &  &  & 31 & 48.9 & 66.8  \\
    RT-DETR-R34 & $\checkmark$ &  &  & 33.4 & 52.1($\uparrow$\textbf{3.2}) & 70.4($\uparrow$\textbf{3.6})  \\
    RT-DETR-R34 &  & $\checkmark$ &  & 31.7 & 50.2 ($\uparrow$\textbf{1.3}) & 68.7($\uparrow$\textbf{1.9})  \\
    RT-DETR-R34 &  &  & $\checkmark$ & 32.1 & 49.9($\uparrow$1.0)  &  68.6($\uparrow$1.8) \\
    RT-DETR-R34 & $\checkmark$ &  & $\checkmark$  & 34.5 & 53.5($\uparrow$4.6) & 70.6 ($\uparrow$3.8) \\
    RT-DETR-R34 & $\checkmark$ & $\checkmark$ &  & 34.1 & \textbf{54.6}($\uparrow$\textbf{5.7}) & \textbf{71.6}($\uparrow$\textbf{4.8})  \\
    \midrule
    RT-DETR-R50 &  &  &  & 42 & 53.1 & 71.3  \\
    RT-DETR-R50 & $\checkmark$ &  &  & 44.4 & 54.4($\uparrow$\textbf{1.3}) & 73.1($\uparrow$\textbf{1.8})  \\
    RT-DETR-R50 &  & $\checkmark$ &  & 42.7 & 53.9($\uparrow$\textbf{0.8}) & 72.3($\uparrow$\textbf{1.0}) \\
    RT-DETR-R50 &  &  & $\checkmark$ & 43.1 & 53.5($\uparrow$0.4) & 72.0($\uparrow$0.7)  \\
    RT-DETR-R50 & $\checkmark$ &  & $\checkmark$ & 45.5 & 54.7($\uparrow$1.6) & 72.6($\uparrow$1.3) \\
    RT-DETR-R50 & $\checkmark$ & $\checkmark$ &  & 45.1 & \textbf{55.7}($\uparrow$\textbf{2.6}) & \textbf{74.0}($\uparrow$\textbf{2.7})  \\
    \bottomrule
  \end{tabular}
\end{table*}

\section{Experimental Results}
\subsection{Experimental Settings}
All experiments were conducted on a workstation equipped with one NVIDIA A100 GPU with 80 GB of VRAM and an Intel Xeon CPU with 128 GB of RAM. We train the Association-DETR-R34 and the Association-DETR-R50 in same configurations. More specifically, we use AdamW\cite{loshchilov2017decoupled} as optimizer, the base learning rate is $1e-4$, the training epoch is $72$, the batch-size for training and evaluation is 16, the number of Hybrid Encoder layer is $1$, the number of selection query is $300$, the number of Background Attention Module and Association Module are $1$, the embedding dims is $256$. Most of our training parameters are same as our baseline RT-DETR\cite{rt-detr}. The hyper-parameter details is presented in Table \ref{parameters}. Moreover, the hyper-parameters for pre-training the Background Attention Module on the Stanford Background Dataset is same parameters as \cite{targ2016resnet}.

\subsection{Dataset}
\subsubsection{Stanford Background Dataset}
The Stanford Background Dataset was introduced by Gould et al. \cite{gould2009decomposing} to evaluate methods for geometric and semantic scene understanding. It contains total 715 images selected from public datasets and the selection criteria required images to depict outdoor scenes, contain at least one foreground object, and include a visible horizon within the image. Semantic and geometric labels were obtained using Amazon Mechanical Turk. Moreover, there are total 9 kind of background class. They are sky, tree, road, grass, water, building, mountain, foreground and unknown respectively.

\subsubsection{COCO}
The COCO dataset is a widely used benchmark for computer vision tasks like object detection, instance segmentation, keypoint detection, and image captioning. It contains complex everyday scenes with multiple objects and rich context, serving as a standard for evaluating visual recognition systems in realistic settings.The COCO val2017 subset, the primary validation split of the COCO 2017 release, contains 5,000 images from the full dataset. It includes 80 object categories for detection and segmentation, with each image annotated with precise instance masks, bounding boxes, category labels, and keypoints for human pose estimation when applicable. Annotations were collected through extensive crowdsourcing to ensure accuracy and consistency. COCO val2017, featuring a balanced mix of diverse object classes and scenes, is widely used for model evaluation and hyperparameter tuning. It provides a standardized benchmark to assess model generalization before test server submission. Its detailed annotations and visual complexity are crucial for developing and fairly comparing state-of-the-art computer vision models.
\subsection{SOTA comparison}
Table \ref{sota_comparison_table} compares our Association DETR with current real-time detectors (YOLO models) and end-to-end detectors (DETR models). Specifically, we compare the M, L, and X variants of YOLOv10, YOLOv11, and YOLOv12, as all achieve outstanding performance on the COCO dataset. Our Association DETR and the YOLO detectors use a common input size of 640 × 640 pixels, while other DETR models use an input size of 800 × 1333 pixels. The FPS measurements were obtained on an NVIDIA T4 GPU using TensorRT FP16 precision. All YOLO detectors employed official pre-trained models available on their respective websites. Our Association DETR-R34 achieves 54.60 in AP$^{val}$, 71.6 in AP$^{val}_{50}$ and 153 FPS, representing the best performance among models with fewer than 40 million parameters. Meanwhile, our Association DETR-R50 achieves 55.7 in AP$^{val}$, 74.0 in AP$^{val}_{50}$ and 104 FPS. Both Association DETR-R34 and Association DETR-R50 outperform other YOLO detectors and DETR models with similar scale.

Moreover, to evaluate the effectiveness of our proposed Association Module, we integrate our Association Encoder (AE) into various models, including RT-DETR-v2-R50, RT-DETRv2-R34 \cite{rtdetrv2}, DETR \cite{carion2020endtoendobjectdetectiontransformers}, and Deformable DETR \cite{zhu2020deformable}. The experimental results are presented in Table \ref{effectiveness_table}. As shown, our Association Module incurs a slight reduction in speed but achieves higher performance compared to each baseline model. Specifically, for the RT-DETR-R34 model, it increases AP$^{val}$ by 5.7 and AP$^{val}_{50}$ by 4.8, while reducing FPS by less than 5.7\%. Similarly, it improves AP$^{val}$ by 2.6 and AP$^{val}_{50}$ by 2.7 for RT-DETR-R50.  Moreover, 2.5 in AP$^{val}$ and 2.7 in AP$^{val}_{50}$ for RT-DETRv2-M; 0.3 in AP$^{val}$ and  0.6 in AP$^{val}_{50}$ for RT-DETRv2-L with R50 as backbone; DETR-R50, it increases 2.7 in AP$^{val}$ and 3.6 in AP$^{val}_{50}$. We need to noticed that it even outperforms than DETR-R101(base). Moreover, 2.6 in AP$^{val}$ and 3.2 in AP$^{val}_{50}$ for Deformable DETR. All these models outperform their respective baselines after integrating our Association Encoder (AE). Notably, the Association-DETR-R34 and Association-DETR-R50 models achieve state-of-the-art performance compared to other models. These experimental results indicate that the AE is a lightweight plug-in module that enhances model performance by incorporating background information.
\subsection{Ablation}
\label{sec:ablation}
For evaluating the importance and effectiveness of each module and comparing them with the basic components (basic vistion transforemr encoder layer), we conduct ablation experiments. The experimental results are presented in Table \ref{ablation-table}. We streamline each module for of Association Encoder in the Association DETR, you can tell both of these modules only consists of 3.1 million parameters according to the Table \ref{ablation-table}. More specifically, the BAM (Background Attention Module) is implemented with only 2.4 million parameters, while the AM (Association Module) uses only 0.7 million parameters. The EL (basic vision transformer encoder layer) with multi-head attention (8 heads) is implemented with 1.1 million parameters. Furthermore, we observe that both BAM and AM improve the baseline model. However, BAM contributes an increase of 3.2 in $AP^{val}$ and 3.6 in $AP^{val}_{50}$ for the RT-DETR-R34 model, and an increase of 1.3 in $AP^{val}$ and 1.8 in $AP^{val}_{50}$ in RT-DETR-R50. Similarly, The AM contributes 1.3 in $AP^{val}$ and 1.9 in $AP^{val}_{50}$ for the RT-DETR-R34 model, and 0.8 in $AP^{val}$ and 1.0 in $AP^{val}_{50}$ in another setting. Given that AM is implemented with only 0.7 million parameters, it is reasonable that it yields a smaller performance boost than BAM. Moreover, although EL has 0.4 million more parameters than AM, it performs worse both when compared to AM alone and when combined with BAM. These results indicate that our proposed BAM and AM modules are effective and offer a good trade-off between performance and speed.
\begin{table}[ht]
\caption{hyperparameters of Association DETR. Most of hyper-parameters are as same as our baseline RT-DETR}
\label{parameters}
\begin{center}
\begin{tabular}{cc}
\toprule
Item & Value \\
\toprule
optimizer & AdamW\\
base learning rate &  1e-4\\
learning rate of backbone &  1e-5\\
freezing BN &            True\\
linear warm-up start factor   &  0.001\\
linear warm-up steps & 2000\\
weight decay & 0.0001\\
clip gradient norm & 0.1\\
ema decay & 0.9999 \\
number of encoder layers & 1\\
number of Max Pooling layers & 1\\
number of input channels $C_i$ & 256 \\
number of reduced channels $C_j$ & 32\\
feedforward dim & 1024\\
number of selected queries & 300\\
number of Assocaition Encoder Layer & 1\\
number of Background Attention Module & 1\\
number of Association Module & 1\\
number of decoder layers & 1 \\
max object detections & 100 \\
epoch & 72 \\
batch-size & 8 \\
% Distance Only th=2.0 & 48.18 & 47.32\\
% Distance Only th=3.0 & 46.27 & 47.37\\
\hline
\end{tabular}
\end{center}
\end{table}
\section{Conclusion}
In this paper, for bridging the gap that no object detection model explicitly utilizes background information and just let it slip away, we propose Association DETR, which leverages background information and achieves state-of-the-art performance on the COCO dataset with 55.7 mAP. Moreover, the proposed Association Encoder (AE) is a lightweight plug-in module that can be integrated into any existing DETR model, enhancing its performance while sacrificing minimal speed.

{
    \small
    \bibliographystyle{ieeenat_fullname}
    \bibliography{main}
}


\end{document}